\documentclass[letterpaper]{article} 
\usepackage{aaai2026}  
\usepackage{times}  
\usepackage{helvet}  
\usepackage{courier}  
\usepackage[hyphens]{url}  
\usepackage{graphicx} 
\usepackage{natbib}  
\usepackage{caption} 
\usepackage{algorithm}
\usepackage{algorithmic}
\usepackage{algorithm}
\usepackage{algorithmic}
\usepackage{multicol}
\usepackage{multirow}
\usepackage{booktabs}
\usepackage{amsmath}
\usepackage{newfloat}
\usepackage{listings}
\DeclareCaptionStyle{ruled}{labelfont=normalfont,labelsep=colon,strut=off} 
\floatstyle{ruled}
\newfloat{listing}{tb}{lst}{}
\floatname{listing}{Listing}
\title{TinyChemVL: Advancing Chemical Vision-Language Models via Efficient Visual Token Reduction and Complex Reaction Tasks}
\author{
   Xuanle Zhao\textsuperscript{\rm 1,2}, Shuxin Zeng\textsuperscript{\rm 1,2}, Xinyuan Cai\textsuperscript{\rm 1}\thanks{Corresponding Authors.}, Xiang Cheng\textsuperscript{\rm 1},\\
    Duzhen Zhang\textsuperscript{\rm 1}, Xiuyi Chen\textsuperscript{\rm 1}, Bo Xu\textsuperscript{\rm 1,2}\footnotemark[\value{footnote}]
}
\affiliations{

     \textsuperscript{\rm 1} The Key Laboratory of Cognition and Decision Intelligence for Complex Systems,\\ Institute of Automation, Chinese Academy of Sciences \\
     \textsuperscript{\rm 2} School of Artificial Intelligence, University of Chinese Academy of Sciences, Beijing, China \\
}

\begin{document}

\maketitle

\begin{abstract}
While Vision Language Models (VLMs) have demonstrated remarkable capabilities in general visual understanding, their application in the chemical domain has been limited, with previous works predominantly focusing on text and thus overlooking critical visual information, such as molecular structures.
Current approaches that directly adopt standard VLMs for chemical tasks suffer from two primary issues: (i) computational inefficiency of processing entire chemical images with non-informative backgrounds. (ii) a narrow scope on molecular-level tasks that restricts progress in chemical reasoning.
In this work, we propose \textbf{TinyChemVL}, an efficient and powerful chemical VLM that leverages visual token reduction and reaction-level tasks to improve model efficiency and reasoning capacity. Also, we propose \textbf{ChemRxn-V}, a reaction-level benchmark for assessing vision-based reaction recognition and prediction tasks. Directly predicting reaction products from molecular images poses a non-trivial challenge, as it requires models to integrate both recognition and reasoning capacities. Our results demonstrate that, with only 4B parameters, TinyChemVL achieves superior performance on both molecular and reaction tasks, while also demonstrating faster inference and training speeds compared to existing models. Notably, TinyChemVL outperforms ChemVLM while utilizing only 1/16th of the visual tokens. This work builds efficient yet powerful VLMs for chemical domains by co-designing model architecture and task complexity.
\end{abstract}

\begin{links}
    \link{Code}{https://github.com/xxlllz/TinyChemVL}
\end{links}

\section{Introduction}
The remarkable success of Large Language Models (LLMs) on diverse tasks, from question answering to code generation \cite{zhang2024internlm, grattafiori2024llama}, has spurred significant interest in leveraging their reasoning capabilities to accelerate scientific discovery. This trend is particularly prominent in domains like mathematics and chemistry, where core entities such as equations and molecules can be parsed in structured textual formats \cite{lu2022dynamic, lightman2023let, zhang2024chemllm, zhang2025scientific}. Especially in the chemical field, the molecules are parsed into SMILES \cite{lightman2023let} and SELFIES \cite{krenn2022selfies} formats for the following analysis. However, representing the chemical molecules in text format inevitably omits some spatial information.
More recently, Vision Language Models (VLMs) have demonstrated remarkable capabilities in addressing a wide range of visual tasks, such as image captioning and question answering \cite{liu2023llava, zhao2025chartcoder}, which can handle more complex tasks. However, in the chemical field, there is still a lack of exploration on how to efficiently utilise VLMs for chemical tasks. Recent works have demonstrated initial attempts in adapting VLMs for chemical applications such as molecule caption and property prediction \cite{li2025chemvlm}. However, the models are generally fine-tuned on general domain VLMs, which limits their efficiency in handling chemical problems. 


Although previous works have made some initial explorations in the field of vision-based chemistry, several challenges remain to be addressed. (i) \textbf{Visual redundancy in processing molecule image.} In the molecular images, all structural information is contained within the graphical representations of the molecules, while the background region is considered non-informative. Consequently, there is significant redundancy in the visual tokens used to represent the entire image, as most tokens correspond to this uninformative space. Also, the partitioning of the molecular image into patches risks the loss of important structural information. For example, in the ChemVLM \cite{li2025chemvlm}, a $\text{800}\times\text{800}$ image needs 1280 visual tokens for embedding, which is around 100 times the number of tokens in the textual questions, which significantly increases the time and computational cost for both training and inference. 
(ii) \textbf{Limited model performance and evaluation diversity.} Currently, the capacities of chemical VLMs are limited, and evaluations are not comprehensive. Existing works \cite{li2025chemvlm, tan2025chemmllm} focus on tasks at the molecular level, such as SMILES Optical Character Recognition (OCR) and property prediction, while ignoring issues at the reaction level. Unlike direct recognition, reaction-level tasks, such as reaction prediction, require the models to perform chemical understanding and reasoning to greater extents.
(iii) \textbf{Bottleneck on molecular image generation.} Recently, ChemMLLM \cite{tan2025chemmllm} proposes that generating the molecular image would expand the applications of VLMs in chemistry and combine VQ-GAN \cite{esser2021taming} with LLMs to generate images. However, purely generating molecular images requires external tools to parse images back into SMILES strings for subsequent analysis. Also, there is a mismatch between the architecture of current VLMs and the requirements of direct image generation.

In this work, we propose TinyChemVL, a small-scale VLM with only 4B parameters, achieving the state-of-the-art (SOTA) performance across various vision-based chemical tasks. (i) To reduce visual token redundancy, we adopt an adaptive token merge and pruning approach \cite{bolya2022token, zeng2024m2m}. Utilizing the same InternVL architecture, our model reduces visual tokens to 1/16 of those in ChemVLM, significantly decreasing redundancy and capturing complex molecular structure.
(ii) To advance the field of vision-based chemistry, we introduce a new benchmark named VisRxnBench. It is designed to evaluate vision-based reaction recognition and prediction tasks, providing 5,000 samples for each. To the best of our knowledge, we are the first to define the task of predicting products directly from the reactant image, which expands the scope of vision-based chemistry.
Furthermore, we conduct a large-scale and diverse dataset for training. Benefiting from the reduced visual tokens, training time on large-scale data increases negligibly.
(iii) To enable TinyChemVL to generate the molecular images. We employ executable code as the intermediary content for generating molecular images \cite{zhao2025vincicoder}. This approach streamlines the process because the SMILES string is directly included within the generated code, enabling the generation molecular images and eliminating the need for additional tools in subsequent analyses.
We evaluate TinyChemVL on various benchmarks, and the results show that it achieves the SOTA performance on most tasks with only 4 billion parameters.
Beyond its strong performance, TinyChemVL exhibits significantly reduced training and inference times compared to existing models. In summary, the main contributions of this work are as follows:
\begin{itemize}
    \item We introduce TinyChemVL, an efficient chemical VLM that achieves SOTA performance in vision-based chemical tasks, with only 4B parameters.
    \item We adopt a token merging and pruning strategy to reduce visual tokens from large background regions and merge complex structures in molecular images. This approach significantly reduces the number of visual tokens, leading to substantially faster inference and training speeds.
    \item We introduce VisRxnBench, a novel benchmark alongside a large-scale training dataset for evaluating vision-based, reaction-level tasks. Our evaluation shows existing VLMs perform poorly on reaction-level tasks, especially when directly predicting products from images.
\end{itemize}

\section{Related Works}
\subsection{Chemical Language Models} Leveraging language models to solve chemical-related tasks has garnered significant attention from the research community. Previous works \cite{edwards2021text2mol, edwards2022translation} utilize pretrained SciBERT \cite{beltagy2019scibert} and T5 \cite{raffel2020exploring} as the backbones and fine-tune on molecule-to-natural language translation tasks. However, these works are generally task-specific. Building on the powerful generalization capabilities of LLMs, chemical LLMs demonstrate improved generalization across a wider range of chemical tasks. Recent works such as ChemLLM \cite{zhang2024chemllm} and ChemDFM \cite{zhao2024chemdfm} achieve this by conducting large-scale chemical instruction tuning datasets and fine-tuning on existing LLMs such as InternLM. UniMoT \cite{zhang2024unimot} utilize a molecular tokenizer instead of a projection layer to encode the molecular graph information. ChemCrow \cite{m2024augmenting} integrates expert-designed tools with GPT-4 as the backbone for chemistry tasks.

\subsection{Vision Language Models}
Recently, leveraging the projection layer to align LLMs and vision encoders, VLMs have achieved superior performance on many visual tasks \cite{zhang2024mm, bi2024visual}. Current models, like GPT-4o \cite{openai2024gpt4o} and InternVL \cite{chen2024expanding} have shown strong performance on vision understanding and reasoning tasks. Recently, VLMs have been increasingly applied to vision-based tasks in the field of chemistry. For instance, ChemVLM \cite{li2025chemvlm} introduces vision-based chemical datasets and utilizes InternVL architecture \cite{chen2024expanding} as the backbone for fine-tuning. ChemDFM-X \cite{zhao2024chemdfmx} utilizes distinct encoders to align representations from visual and graph modalities with the texts. In the generative domain. ChemMLLM \cite{zhang2024chemllm} employs the VQ-GAN \cite{esser2021taming} as the encoder and decoder for molecular image understanding and generation. 
However, these models are typically directly fine-tuned on existing architectures and suffer from significant visual token redundancy, resulting in computational inefficiency. To address these issues, we propose TinyChemVL, which employs an adaptive token merging and pruning strategy, along with a large-scale and diverse training corpus, to enhance both model efficiency and performance.


\section{Methods}
\begin{figure*}[t]
\centering
\includegraphics[width=0.95\textwidth]{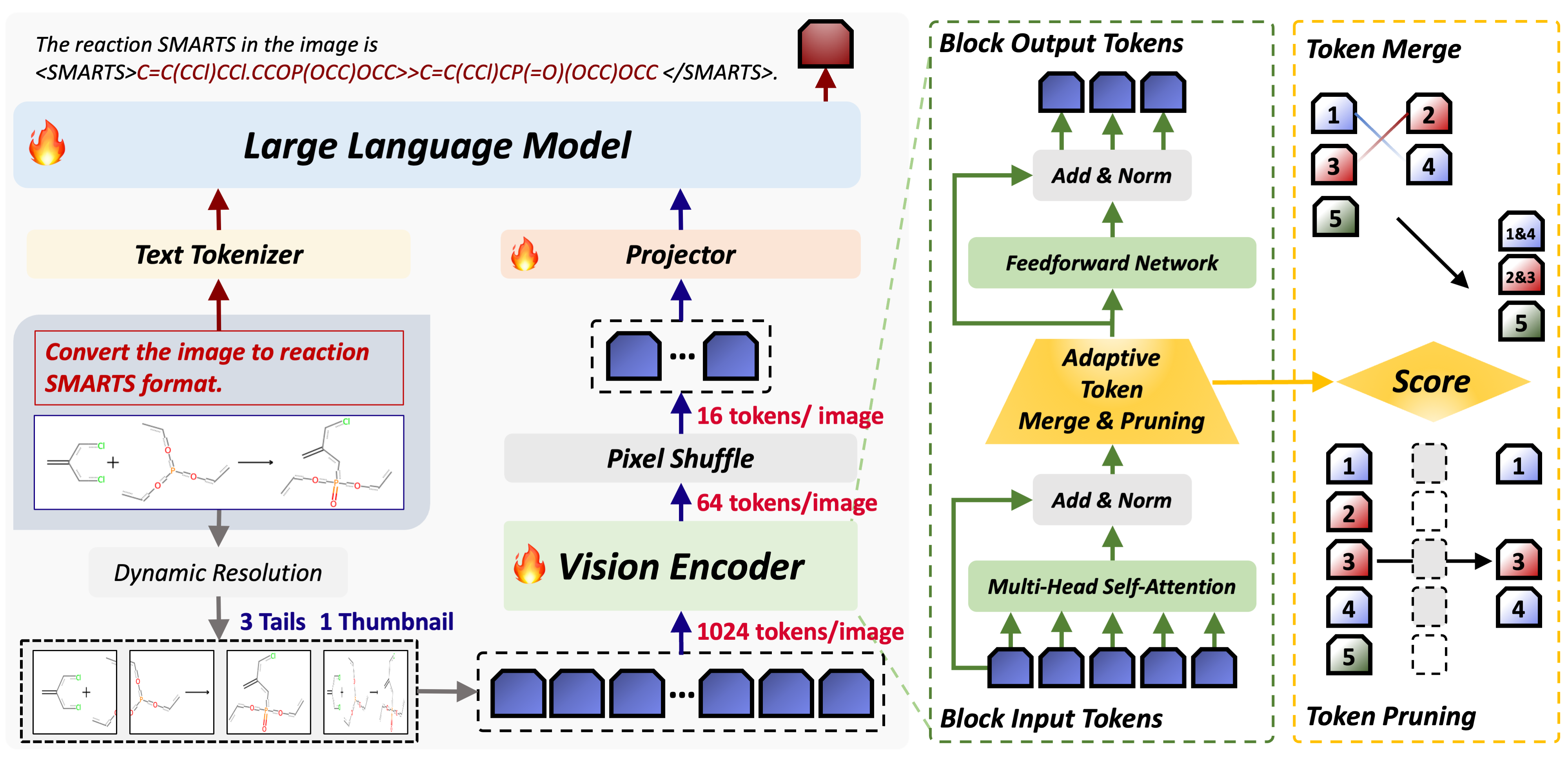} 
\vspace{-5pt}
\caption{The overview structure of TinyChemVL. For the input image, a dynamic resolution strategy is employed to segment and downsample it to standard resolutions. The vision encoder progressively reduces the number of visual tokens through the adaptive token merging and pruning method. The pixel shuffle operation is the default strategy utilized in InternVL2.5.}
\label{fig:main}
\vspace{-5pt}
\end{figure*}

The overall structure of TinyChemVL is illustrated in Figure~\ref{fig:main}. TinyChemVL adopts the well-established ViT-MLP-LLM architecture, following the same framework conducted by InternVL \cite{chen2024expanding} and ChemVLM \cite{li2025chemvlm} to ensure fair comparisons. In contrast to prior works, we adopt InternVL2.5-4B as our backbone. To process high-resolution images, the dynamic resolution strategy segments the image into several $448\times448$ tiles based on its dimensions and concatenates them with the downsampled image.

\subsection{Visual Token Redundancy}
The vision transformer aims to encode the image into visual features. Given the input image resolution $N\times N$, the ViT encode the image to $(N//P)^2$ tokens with $P\times P$ patch size. When facing the high-resolution image, the dynamic resolution strategy first segments it into standard-resolution tiles, followed by a directly down-sampled thumbnail. Then the ViT encode each tile independently and concatenates the processed tokens to conduct the visual tokens. We identify a severe modality imbalance when representing chemical molecular images. A standard-resolution (448$\times$448) molecular image is encoded into 256 visual tokens, approximately 8.5 times more than the around 30 textual tokens required for its corresponding SMILES representation. (Based on an average of 200k SMILES embedded with the InternVL2.5 tokenizer). As molecular images are characterized by inherent sparsity, most visual tokens correspond to blank backgrounds. This results in significant visual token redundancy, imposing a substantial computational burden on chemical VLMs.
Furthermore, the patching process disrupts molecular structures by fragmenting the visual representation. To mitigate the above challenges, we introduce an adaptive token merging and pruning strategy.

\subsection{Adaptive Token Merge and Pruning}
The visual token reduction strategy, positioned between the attention and FFN layers in each transformer block, adaptively prunes inattentive tokens and merges duplicative ones based on the current visual token distribution \cite{bolya2022token, kim2024token}.
In detail, token scores are first calculated to quantify the task-specific importance of each token. While numerous non-parametric token scoring mechanisms have been proposed  \cite{liang2022not, xu2022evo, zeng2025token}, we adopt the Adaptive Token Sampler (ATS) \cite{fayyaz2022adaptive} specifically for its ability to incorporate the value matrix $\boldsymbol{V}$ in calculations.
\begin{gather}
\boldsymbol{A}=\textit{Softmax}\left(\frac{\boldsymbol{Q} \cdot \boldsymbol{K}^T}{\sqrt{d}}\right) \\
\text{Score}_i=\frac{\boldsymbol{A}_{1, i+1} \times\left\|\boldsymbol{V}_{i+1}\right\|}{\sum_{j=1}^N \boldsymbol{A}_{1, j+1} \times\left\|\boldsymbol{V}_{j+1}\right\|},
\end{gather} 
where $\boldsymbol{Q}$, $\boldsymbol{K}$, $\boldsymbol{V}$ are the query, key, values of self-attention. $\boldsymbol{A}$ is the attention metric. $\text{Score}_i$ is the importance score assigned to image tokens.

\textbf{Token Pruning} For token pruning, we adopt the standard Top-K selection strategy \cite{wu2023ppt, liang2022not}, which preserves a fixed number (K) of the highest-scoring tokens. Since the total number of input tokens at each layer is pre-defined, this approach is equivalent to pruning a fixed number of tokens, which provides predictable control over the compression ratio.

\textbf{Token Merge} For token merging, we leverage the Bipartite Soft Matching (BSM) algorithm \cite{bolya2022token} due to its superior performance.
BSM first partitions the tokens into two equal-sized sets. It then constructs a bipartite graph by creating an edge between each token in one set and its most similar counterpart in the other, using cosine similarity as the metric. After selecting the Top-K edges with the highest similarity scores, the remaining connected tokens are merged via a weighted average of their features. Crucially, this merging process is not constrained by spatial adjacency. Non-adjacent tokens can also be merged if they belong to different subsets and are similar enough.

\textbf{Proportional Attention}
As the sequence of tokens is condensed by merging and pruning after each attention block, we follow ToMe \cite{bolya2022token} to maintain a row vector $s$ that tracks the number of original tokens each current token represents, thereby preserving information fidelity. This row vector is also used as a weight to reflect its importance in the calculation of the attention matrix
\begin{equation}
\boldsymbol{A}=\operatorname{Softmax}\left(\frac{\boldsymbol{Q} \cdot \boldsymbol{K}^T}{\sqrt{d}}+\log \mathbf{s}\right) .
\end{equation}

\textbf{Adaptive Policy} Consistent with trends observed in prior work \cite{wu2023ppt}, our analysis of the InternViT-300M-448px-V2.5 encoder confirms that the variance of token significance scores for a given sample correlates positively with model depth. This inspired us to employ an adaptive strategy to mitigate the token redundancy. 
When token significance scores exhibit low variance, it indicates a convergence of importance, often corresponding to a large background area. In this situation, these uninformative tokens are pruned. Conversely, when the scores show high variance, it signifies that some tokens are much more important than others and likely represent complex structures. In these cases, tokens corresponding to adjacent or functionally similar molecular structures are merged.
This approach reduces the number of visual features by removing irrelevant tokens, while simultaneously creating a more efficient and holistic structural representation of the molecule.
Furthermore, since the variance of token significance scores differs across instances even within the same layer, we propose a strategy that adapts at both the instance and layer levels.
\begin{align}
    S_{op_i} &= \textit{var}(\text{Score}_i), \\
    \text{Policy}_i &= 
        \begin{cases}
            \text{Token Pruning,} & \text{if } S_{op_i} \le \tau \\
            \text{Token Merge,} & \text{otherwise}
        \end{cases}
\end{align}
where $\tau$ is a decision threshold hyperparameter that balances the trade-off between the pruning and merge strategies, which we set to a default value of $1e-5$ based on a statistical analysis of score variances across existing chemical image datasets.

\section{Data Construction}
\subsection{Training Data}
A primary challenge in training effective VLMs for chemistry is the curation of suitable training datasets. While numerous chemical datasets have been collected, the majority of them are in text \cite{zhang2024chemllm, fang2023mol} or graph \cite{park2024llamo, lee2025mol} formats, with a notable lack of image data. However, training high-performing VLMs necessitates a large-scale and diverse image-text corpus. As the training corpus of ChemVLM \cite{li2025chemvlm} is not open-source, we leverage the training corpora provided by the more recent work ChemMLLM \cite{tan2025chemmllm} as the initial datasets. However, we find that the scope and diversity of this training dataset are still limited. Therefore, we construct a large-scale training dataset containing both molecular and reaction-level tasks.

\textbf{Molecule Recognition} For vision-based chemical tasks, where molecules are represented as images, the capacity to recognize their structures is a foundational skill. This task requires the VLM to parse the visual representation of a molecule into its SMILES format, thereby unlocking subsequent chemical analyses. We collect training corpus from ChEBI-20-MM \cite{liu2025quantitative}, MolGrapher \cite{morin2023molgrapher}, MolScribe \cite{qian2023molscribe}, ORDerly \cite{wigh2024orderly} and EDU-CHEMC \cite{chang2025rfl}. For data already available as $<$image, SMILES$>$ pairs, we incorporate them directly into our training corpus. For all other instances, which consist solely of SMILES strings, we generate the corresponding molecular images using RDKit \cite{bento2020open} and Indigo \cite{epam_indigo_2024}. To ensure visual diversity, we manually define a diverse range of seed drawing parameters, including color, size and style, which are then randomly combined and applied during image rendering. Furthermore, our experiments reveal that recognizing SMILES strings from handwritten molecular images poses a significant challenge. To address this, we adapt the EDU-CHEMC dataset \cite{chang2025rfl} by developing a parser code that converts its native structure-specific markup language (SSML) \cite{hu2023handwritten} into corresponding SMILES, yielding a paired dataset of handwritten images and their corresponding SMILES string.

\textbf{Reaction Recognition.}
Beyond individual molecules, we also apply TinyChemVL to the task of whole reaction recognition, which is foundational to understanding chemical reactions. To address the scarcity of public data on chemical reactions, we utilize the same method denoted in molecular image generation and render reaction images from the ORDerly dataset \cite{wigh2024orderly}. For each input image, the task is to parse the depicted chemical reaction into the reactants$>$reagents.solvents$>$products format. Each component, expressed in SMILES format, may represent more than one molecule. In cases with multiple molecules, their respective SMILES strings are concatenated and separated by a period (e.g., c1ccccc1.Cl).

\textbf{Property Prediction.}
Following the property prediction task of ChemMLLM, we construct training data involving first decomposing reactions in the ORDerly dataset into molecules, then sampling SMILES based on their length distribution, and finally utilizing RDKit to calculate molecular properties. The properties include molecular weight (MW), logarithm of the Partition Coefficient of a solute between octanol and water (LogP), Topological Polar Surface Area (TPSA), Hydrogen Bond Donor (HBD), Hydrogen Bond Acceptor (HBA), Rotatable Bond (RB), and Quantitative Estimate of Drug-likeness (QED). 

\textbf{Reaction Prediction.}
Drawing inspiration from chemical experts who can predict reaction products simply by observing the structures of the reactants, we construct a vision-based reaction prediction dataset to train TinyChemVL for this task. Specifically, the model learns to predict the product from an image depicting the reactants, reagents, and solvents. 
Additionally, to better assist the model in understanding the input image, we also construct the training corpus that pairs the input images with the corresponding SMILES strings. All the data for this task is sampled from the ORDerly dataset as well.

\textbf{Molecular Image Generation.}
ChemMLLM introduces the task of conditional chemical image generation, which requires models to generate a target molecular image based on specified properties or optimization requirements. However, this task remains text-centric, as performance evaluation ultimately relies on converting the generated image back into a SMILES string using tools like MolScribe \cite{qian2023molscribe}. In contrast, we propose a more direct and verifiable paradigm that utilizes executable \texttt{Python} code to render the target molecular image. To implement this, we sample and reconstruct the ChemMLLM training data by replacing its \texttt{<|image|>} tokens with the corresponding rendering code. We also collect further property-to-image data to expand the training corpus.

Table~\ref{tab:data} demonstrates the composition of our training dataset for each task, distinguishing between pre-existing samples and those we construct.

\begin{figure}[t]
\centering
\includegraphics[width=0.48\textwidth]{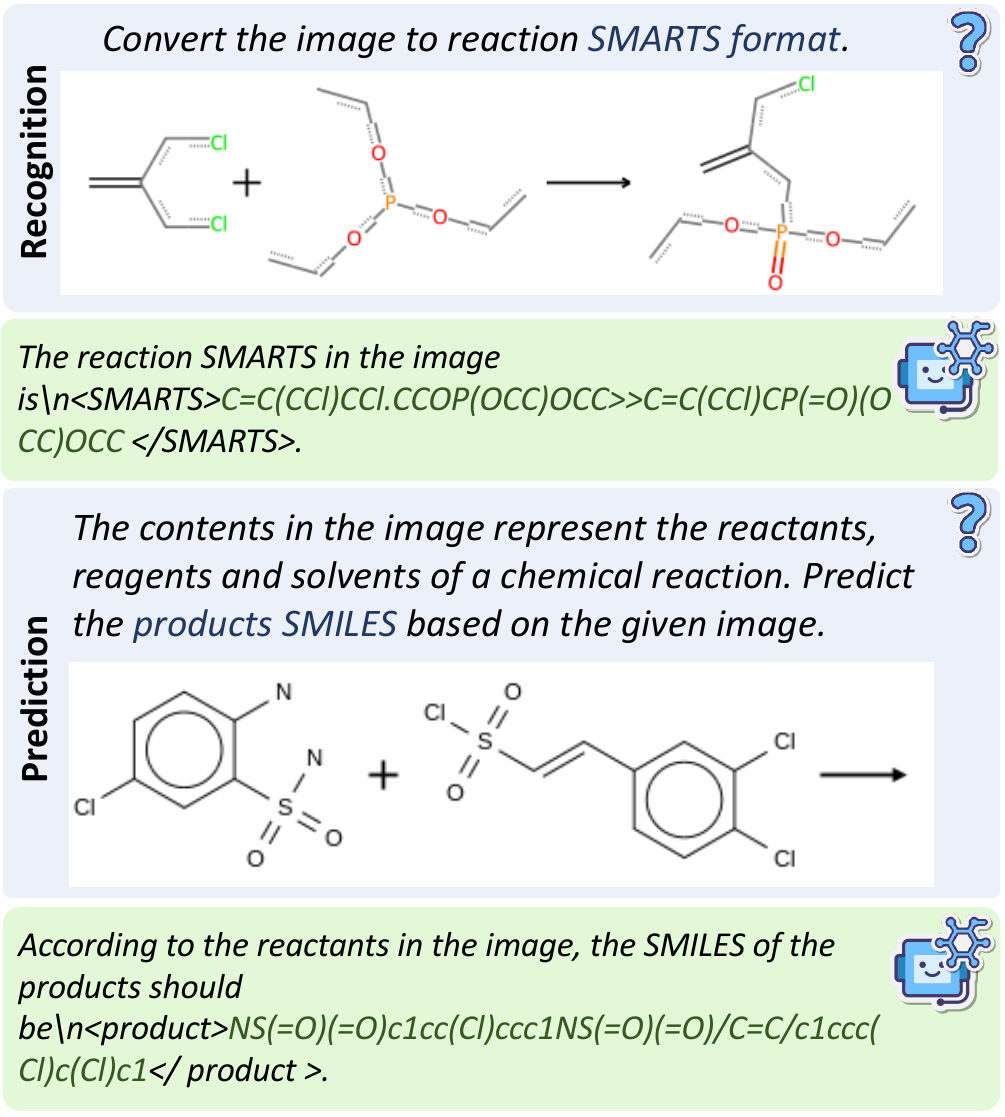} 
\caption{The demonstration about vision-based reaction recognition and prediction tasks in ChemRxn-V.}
\label{fig:reaction}    
\end{figure}

\begin{table}[t]
\centering
\begin{tabular}{lcc}
\toprule
\textbf{Task} & \textbf{Total} & \textbf{Ours}\\
\midrule
Molecule Recognition & $500k$ & $385k$\\
Reaction Recognition & $200k$ & $200k$ \\
Property Prediction & $150k$ & $55k$\\
Reaction Prediction & $200k$ & $200k$ \\
Molecular Image Generation & $200k$ & $55k$ \\
\bottomrule 
\end{tabular}
\caption{Quantity of training samples across different tasks, showing the total number of samples (\textbf{Total}) and the subset generated by ourselves (\textbf{Ours}). The remaining samples are collected from existing vision-based chemical datasets.}
\label{tab:data}
\end{table}

\subsection{Evaluation}
For the vision-based reaction recognition and prediction tasks, we introduce a new evaluation benchmark named (\textbf{ChemRxn-V}). This benchmark, containing reaction recognition and prediction tasks, is curated by filtering the ORDerly test set \cite{wigh2024orderly} and subsequently rendering the corresponding reaction and reactant images using RDKit and Indigo. Specifically, each task in the ChemRxn-V benchmark contains 5,000 test samples, which are selected based on reaction length to ensure the distribution is representative of the overall dataset.
Prediction performance is assessed by computing the RDKit fingerprint similarity between predicted and ground-truth products, following the Mol-instructions \cite{fang2023mol} protocol. In contrast, evaluating recognition performance requires a more detailed approach, as RDKit cannot directly compare reactions. Therefore, we first decompose each reaction into its three components (reactants, reagents/solvents, products), then calculate the fingerprint similarity for each part, and finally compute a weighted average score based on the molecule count per component.

\subsection{Training Details}
We utilize InternVL2.5-4B as the backbone model for supervised fine-tuning (SFT). The visual token reduction strategy provides the computational efficiency necessary for full-parameter fine-tuning. The model is trained on 8 NVIDIA A100$\times$80G GPUs for 1.5 epochs. The per-device batch size and the gradient accumulation steps are set to 16 and 2, respectively. All the training process is conducted on ms-swift.

\section{Experiments}
\subsection{Experiment Settings}
\textbf{Baselines} To assess the effectiveness of our proposed TinyChemVL, we compare it with existing models in three setups: (1) General-domain open-source VLMs, including InternVL2.5-4B \cite{chen2024expanding}, Qwen2.5-VL-7B \cite{bai2025qwen2}, LLaVA-v1.5-7B \cite{liu2023llava} and Phi-3.5-vision \cite{abdin2024phi}. (2) Proprietary models include GPT-4V \cite{openai2023gpt4v} and GPT-4o \cite{openai2024gpt4o}. (3) Chemical-domain VLMs including ChemVLM (8B, 26B) \cite{li2025chemvlm}, ChemDFM-X (13B) \cite{zhao2024chemdfmx} and ChemMLLM \cite{tan2025chemmllm}.

\textbf{Benchmarks} To ensure a fair and direct comparison with prior works, we align our evaluation protocols with established methods. For the molecule recognition task, we adopt the benchmarks from ChemVLM (ChemOCR) and ChemMLLM (img2smiles), reporting two primary metrics calculated via RDKit, the average Tanimoto similarity (Avg. Sim.) and the Tanimoto hit at 1.0 (Tani@1.0). 
For property prediction, we also utilize benchmarks in ChemMLLM (img2property) using the Mean Squared Error (MSE) between predicted and ground-truth values for evaluation.
For molecular image generation, we assess performance on the two distinct subtasks proposed in ChemMLLM. For the image-to-image (img2img) task, property improvement is quantified by the increase in LogP values between the input and output molecules. For the property-to-image (property2img) task, we calculate the Mean Squared Error (MSE) between the output and the ground-truth input utilizing RDKit.
Finally, for reaction-level tasks, we evaluate performance using our proposed ChemRxn-V benchmark, utilizing the aforementioned method for evaluation
To evaluate our adaptive visual token reduction strategy, we assess the computational efficiency of TinyChemVL by comparing its inference throughput and training time against two baselines: the general-domain InternVL-2.5-4B and the chemical-domain ChemVLM-8B.

\subsection{Main Results}

\begin{table}[t]
\renewcommand{\arraystretch}{1.02}
\setlength{\tabcolsep}{0.7mm}
\centering
\begin{tabular}{lcccc}
\toprule
\multirow{2}{*}{Models} & \multicolumn{2}{c}{\textbf{ChemOCR (\%)}} & \multicolumn{2}{c}{\textbf{img2smiles(\%)}}\\
\cmidrule(r){2-3} \cmidrule(r){4-5}
& Avg Sim. & Tani\text{@}1.0  & Avg Sim. & Tani\text{@}1.0 \\
\midrule
GPT-4V & $15.0$ & $2.1$ & $9.74$ & $0.01$\\
GPT-4o & $36.8$ & $3.4$ & $29.0$ & $0.01$ \\
Qwen2.5-VL-7B & $25.5$ & $0.4$ & $28.2$ & $0.03$ \\
LLaVA-v1.5-7B  & $9.0$ & $0.0$ & $5.0$ & $0.00$ \\

InternVL2.5-4B & $4.4$ & $0.0$ & $2.0$& $0.01$\\ 
Phi-3.5-vision & $0.4$ & $0.0$  & $1.2$ & $0.01$\\
ChemVLM-26B & $71.0$ & $42.9$ & $47.1$ & $12.1$ \\
ChemVLM-8B  & $\underline{81.7}$ & $\underline{57.7}$ & $55.0$ & $15.0$ \\
ChemMLLM & - & - & $75.0$& $49.0$ \\
ChemDFM-X & $70.9$ & $36.5$ & $\mathbf{90.9}$ & $\mathbf{77.6}$ \\
\midrule
TinyChemVL & $\mathbf{91.2}$ & $\mathbf{77.4}$ & $\underline{89.5}$ & $\underline{75.6}$ \\
\bottomrule 
\end{tabular}
\caption{Evaluation results of the molecular recognition task. The Avg Sim. and Tani\text{@}1.0 are the abbreviations of average tanimoto similarity and tanimoto hit 1.0. The best and second-best performances are indicated in \textbf{bold} and \underline{underline} respectively.}
\label{tab:main_results1}
\end{table}

\begin{table}[t]
\small
\centering
\renewcommand{\arraystretch}{1.02}
\setlength{\tabcolsep}{0.6mm}
\begin{tabular}{lccccccc}
\toprule
\multirow{2}{*}{Models} & \multicolumn{6}{c}{\textbf{img2property $\downarrow$}}\\
 \cmidrule(r){2-8}
 & MW & LogP & TPSA & HBD & HBA & RB & QED \\
\midrule
GPT-4o  & $7.6e3$ & $5.7$ & $1.2e3$ & $1.8$ & $6.2$ & $11.9$ & $0.03$\\
Qwen2.5VL-7B & $5.3e3$ & $22$ & $3.0e3$ & $2.4$ & $12.7$ & $49.4$  & $0.19$\\
LLaVA-v1.5-7B  & $3.0e4$ & $5.1$ & $5.0e4$ &$53.3$ & $23.7$ & $39.9$ & $1.0e5$\\
Phi-3.5-vision & $4.8e4$  & $14.7$ & $2.3e4$ & $3.3$ & $17.7$ & $1.0e6$ & $276.8$  \\
InternVL2.5-4B & $2.3e4$ & $9.7$ & $5.4e3$ & $1.3e4$ & $6.0e4$  & $2.0e4$ & $2.3$ \\ 

ChemVLM-8B  & $1.0e4$ & $4.9$ & $5.0e3$ & $4.2$ & $27.2$ & $45.8$ & $0.24$\\
ChemMLLM  & $\underline{789.7}$ & $\textbf{0.7}$ & $\underline{152.6}$ & $\underline{0.18}$ &$\underline{0.8}$ & $\underline{1.6}$ & $\underline{0.008}$\\
\midrule
TinyChemVL& $\mathbf{488.0}$ & $\underline{0.9}$ & $\mathbf{71.1}$ & $\mathbf{0.09}$ & $\mathbf{0.3}$ & $\mathbf{0.7}$ & $\mathbf{0.003}$\\
\bottomrule 
\end{tabular}
\caption{Evaluation results on the property prediction task. We manually evaluate models that are not reported in the ChemMLLM and calculate the Mean Squared Error (MSE) for comparison.}
\label{tab:main_results2}

\end{table}

\begin{table}[t]
\centering
\renewcommand{\arraystretch}{1.02}
\setlength{\tabcolsep}{0.9mm}
\begin{tabular}{lcccc}
\toprule
\multirow{2}{*}{Models} & \textbf{img2img $\uparrow$} & \multicolumn{3}{c}{\textbf{property2img $\downarrow$}}\\ 
\cmidrule(l){2-2}\cmidrule(l){3-5}
& Increased LogP & MW & LogP & TPSA \\
\midrule
GPT-4o & $1.95$ & $\underline{7633.2}$ & $\underline{5.7}$& $\underline{1209.2}$ \\
LLaVA-v1.5-7B & $-0.86$ & $>2e6$ & $1104.4$ & $>1e5$ \\
ChemLLM-7B & - & $>7e4$ & $10.0$ & $7666.0$ \\
ChemVLM-8B & $0.45$ & $>2e5$ & $9.0$ & $>2e4$ \\
ChemMLLM & $\mathbf{4.2}$ &  $>3e4$ & $13.2$ & $1191.6$ \\
\midrule
TinyChemVL & $\underline{2.2}$ & $\mathbf{1620.2}$ & $\mathbf{0.9}$ & $\mathbf{182.5}$\\
\bottomrule 
\end{tabular}
\caption{Evaluation results on the molecule image generation task. TinyChemVL achieves superior performance on property2img tasks.}
\label{tab:main_results3}
\end{table}

\begin{table}[t]
\centering
\renewcommand{\arraystretch}{1.02}
\setlength{\tabcolsep}{1mm}
\begin{tabular}{lcccc}
\toprule
\multirow{2}{*}{Models} & \multicolumn{2}{c}{\textbf{Recognition}} & \multicolumn{2}{c}{\textbf{Prediction}} \\
\cmidrule(r){2-3}\cmidrule(l){4-5}
& Avg Sim. & EM  & Avg Sim. & Tani\text{@}1.0 \\
\midrule
GPT-4o & $19.1$ & $0.1$ & $\underline{30.4}$ & $\underline{1.4}$ \\
Qwen2.5VL-7B  & $3.8$ & $0.1$ & $11.9$ & $0.0$\\
Phi-3.5-vision & $1.5$ & $0.0$ & $0.8$ &$0.0$ \\
InternVL2.5-4B & $1.6$ &  $0.0$ & $2.7$ & $0.0$\\ 
ChemVLM-8B  & $0.6$ & $0.0$ & $4.8$ & $0.0$\\
ChemDFM-X & $\underline{28.32}$ & $\underline{3.2}$ & $12.7$&  $0.7$\\
\midrule
TinyChemVL  & $\mathbf{93.4}$&  $\mathbf{67.9}$& $\mathbf{78.9}$ & $\mathbf{52.4}$\\
\bottomrule 
\end{tabular}
\caption{The evaluation results on our proposed ChemRxn-V. EM is the abbreviation of Exact Match.}
\label{tab:main_results4}

\end{table}

\textbf{Model Performance.} 
The performance of molecular recognition 
is denoted in the Table \ref{tab:main_results1}. As the foundation task of vision-based chemistry, TinyChemVL achieves SOTA performance on the ChemOCR, significantly surpassing existing VLMs. While ChemDFM-X (13B) slightly outperforms TinyChemVL (4B) on the img2smiles task, our model is significantly more parameter-efficient, achieving nearly equivalent performance.
Furthermore, we compare TinyChemVL with specialized SMILES OCR models, for example, Decimer \cite{rajan2021decimer} reports a 92.6\% average similarity and a 77.3\% Tanimoto@1.0 on the ChemOCR benchmark. TinyChemVL is the first VLM to deliver competitive results with these specialized models. Unlike specialized models that employ specific loss functions for the image-to-SMILES task, our approach demonstrates that general-purpose VLMs have the potential to match their performance. This finding indicates that general VLMs can match the performance of specialist models on specialized tasks without sacrificing their broader and versatile capabilities.

The results for the property prediction tasks are summarized in Table \ref{tab:main_results2}. TinyChemVL demonstrates superior performance across most tasks, achieving approximately half the Mean Squared Error (MSE) of the second-best model, ChemMLLM. The only suboptimal performance is the LogP img2img task.
The performance on the image generation task is denoted in Table \ref{tab:main_results3}. We evaluate baseline models (excluding ChemMLLM) on their direct generation of SMILES strings. In contrast, TinyChemVL generates executable Python code that renders the molecule. For a fair assessment, we extract the SMILES from the generated code for comparison. The results show that TinyChemVL achieve superior performance on the property2img task. In the img2img task, TinyChemVL achieves the second-best performance, surpassed only by the ChemMLLM.

Furthermore, we evaluate model performance on our constructed ChemRxn-V, which contains vision-based reaction recognition and prediction tasks. As shown in Table \ref{tab:main_results4}, TinyChemVL achieves best performance, significantly outperforming existing VLMs that struggle with reaction-level tasks. Notably, for the reaction prediction task, no existing models attempt to generate products based solely on visual representations of the reactants. For chemical experts, the molecular image is the most intuitive medium for understanding reactions and predicting products. Unlike textual notations such as SMILES, which are essentially a human-defined grammar for computation, molecular images offer a more natural representation. Therefore, enabling models to predict reaction products directly from molecular images represents a crucial research frontier. Success in this area is pivotal for establishing VLMs as indispensable tools for chemical research.

Overall, TinyChemVL achieves SOTA performance across a diverse suite of vision-based chemical tasks at both the molecular and reaction levels. Furthermore, we propose ChemRxn-V, a vision-based benchmark for reaction-level tasks that poses a significant challenge to existing models, thereby expanding the scope of this research domain.

\begin{table}[t]
\setlength{\tabcolsep}{1mm}
\centering
\begin{tabular}{lccc}
\toprule
\multirow{2}{*}{Models} & \multicolumn{2}{c}{\textbf{Inference}} & \textbf{Training} \\
\cmidrule(r){2-3} \cmidrule(l){4-4}
 & Sample/s ($\uparrow$) & Avg. Tokens ($\downarrow$) & Hours ($\downarrow$)\\
\midrule
ChemVLM-8B & $7.41$ & $896$ &- \\
InternVL2.5-4B & $9.11$ & $894$ & $47*$\\
\midrule
TinyChemVL & $11.84$ &  $108$ & $15$ \\
\bottomrule 
\end{tabular}
\caption{The results of inference efficiency on ChemOCR. * denotes the estimated time by tqdm. The Avg. Tokens denotes the average number of tokens per input, including both visual and textual tokens.}
\label{tab:main_results5}

\end{table}

\begin{figure}[t]
\centering
\includegraphics[width=0.45\textwidth]{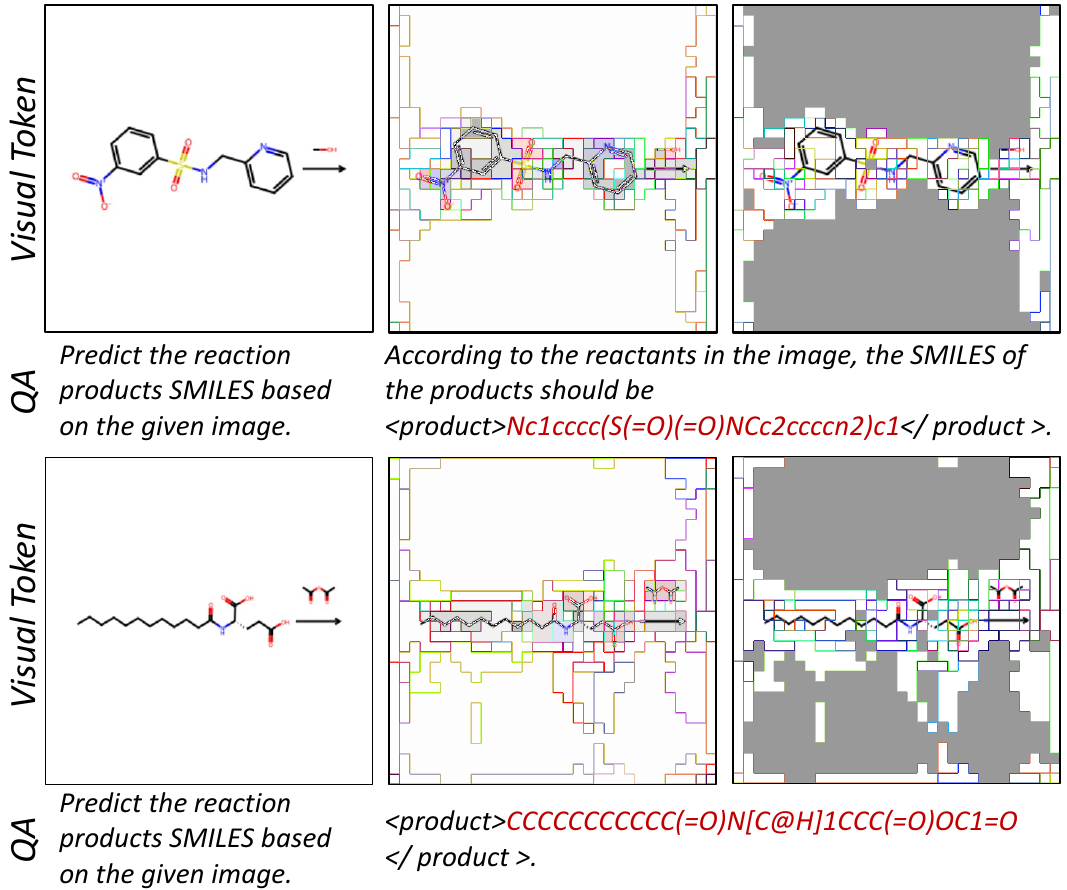} 
\caption{The odd rows (QA) show how the method prunes blank background tokens (indicated by grey areas) and merges molecular structure tokens in a reaction image. The even rows (Visual Tokens) display the associated task instruction and the final answer generated by TinyChemVL.}
\label{fig:redution}
\end{figure}

\textbf{Model Efficiency.}
We conduct two primary comparisons to evaluate our proposed TinyChemVL. First, to specifically assess the effectiveness of our adaptive visual token reduction method, we benchmark TinyChemVL against its backbone model, InternVL2.5-4B. Second, to evaluate the overall efficiency of our approach relative to specialized models, we compare TinyChemVL with ChemVLM-8B, the well-established VLM in the chemical domain. As all these models are based on the InternVL family, we utilize LMDeploy \cite{2023lmdeploy} as the inference toolkit for fairly comparison. We evaluate model efficiency on the ChemOCR dataset by measuring both inference throughput (Samples/s) and computational overhead, which is quantified by the average number of input tokens (Avg. Tokens).
Furthermore, we compare the training time cost of TinyChemVL against InternVL-2.5-4B on our training corpus. To ensure a fair comparison under identical hardware constraints, we run each model at the maximum batch size it can accommodate within the GPU memory budget of A100. This setting reveals that TinyChemVL can utilize a batch size four times larger than InternVL-2.5-4B, forming the basis for our subsequent time measurements.

The results are shown in Table \ref{tab:main_results5}. TinyChemVL achieves the highest inference speed, a result attributed to its fewer input tokens, which requires only about 1/8 input tokens compared to competing models. Notably, TinyChemVL exhibits a significant reduction in training time compared to InternVL2.5-4B, which enhances the efficiency of finetuning and promotes the iteration process.

\begin{table}[t]
\setlength{\tabcolsep}{1mm}
\centering
\begin{tabular}{c|ccc}
\toprule
\textbf{Ablation} & \textbf{ChemOCR} & \multicolumn{2}{c}{\textbf{ChemRxn-V}} \\
 (tokens/image) & Recognition & Recognition & Prediction \\ 

\midrule
16 & $77.4$ & $62.7$ &  $52.4$\\
4 &  $76.2\mathbf{(-1.2)}$ & $59.5\mathbf{(-3.2)}$ & $50.1\mathbf{(-2.3)}$\\
\bottomrule 
\end{tabular}
\caption{The ablation study of different numbers of visual tokens on molecular and reaction tasks. Tani\text{@}1.0 and exact match are utilized for comparison.}
\label{tab:ablation}
\end{table}

\subsection{Ablation Study}
To demonstrate the efficacy of our visual token reduction strategy, Figure~\ref{fig:redution} visualizes the compression results on two chemical reaction images. These results show that our method effectively prunes tokens representing solely the blank background while preserving those crucial for representing the molecules, without influencing the answers.
Also, we conduct an ablation study on the impact of reduced visual token quantity with the same training data. As shown in Table~\ref{tab:ablation}, further reducing visual tokens from 16 to 4 per image results in performance drops, demonstrating the insufficiency of 4 tokens for representing reaction images.

\section{Conclusion}
In this work, we propose TinyChemVL, an efficient and powerful chemical VLM that performs superiorly on vision-based molecular and reaction tasks. By adopting our token reduction strategy, we reduce the number of visual tokens to 1/16th of the original count, ensuring high efficiency for both training and inference. Also, we propose a vision-based reaction-level benchmark, ChemRxn-V, to assess the vision-based reaction recognition and prediction capacities. ChemRxn-V and TinyChemVL could generalize chemical VLMs into more complex scenarios.

\section{Acknowledgements}
This work was supported by
the robotic AI-Scientist platform of Chinese Academy of Sciences.
\bibliography{aaai2026}

\end{document}